\documentclass{article}
\usepackage[a4paper, margin=2cm]{geometry}
\usepackage{authblk}
\usepackage[utf8]{inputenc}
\usepackage{amsmath}
\usepackage{amssymb}
\usepackage{graphicx}
\usepackage{float}
\usepackage{xcolor}

\usepackage[backend=biber,sorting=none,citestyle=numeric-comp]{biblatex}
\title{Detecting Causality in the Frequency Domain with Cross-Mapping Coherence}

\author[1]{Zsigmond Benkő}
\author[1,2]{Bálint Varga}
\author[1]{Marcell Stippinger}
\author[1]{Zoltán Somogyvári}
\affil[1]{HUN-REN Wigner Research Centre for Physics, Department of Computational Sciences, Budapest, Hungary}
\affil[2]{Semmelweis University, Semmelweis Doctoral School, János Szentágothai Neurosciences Division, Budapest, Hungary}

\begin{document}
\maketitle

\begin{abstract}
    Understanding causal relationships within a system is crucial for uncovering its underlying mechanisms.
    Causal discovery methods, which facilitate the construction of such models from time-series data, hold the potential to significantly advance scientific and engineering fields.
    
    This study introduces the Cross-Mapping Coherence (CMC) method, designed to reveal causal connections in the frequency domain between time series. 
    CMC builds upon nonlinear state-space reconstruction and extends the Convergent Cross-Mapping algorithm to the frequency domain by utilizing coherence metrics for evaluation.
    We tested the Cross-Mapping Coherence method using simulations of logistic maps, Lorenz systems, Kuramoto oscillators, and the Wilson-Cowan model of the visual cortex. 
    CMC accurately identified the direction of causal connections in all simulated scenarios.
    When applied to the Wilson-Cowan model, CMC yielded consistent results similar to spectral Granger causality.
    
    Furthermore, CMC exhibits high sensitivity in detecting weak connections, demonstrates sample efficiency, and maintains robustness in the presence of noise.

    In conclusion, the capability to determine directed causal influences across different frequency bands allows CMC to provide valuable insights into the dynamics of complex, nonlinear systems.
        
\end{abstract}
\section{Introduction}

\subsection{Causal discovery from time series}

\begin{figure}[tb!]
    \centering
    \includegraphics[width=\linewidth]{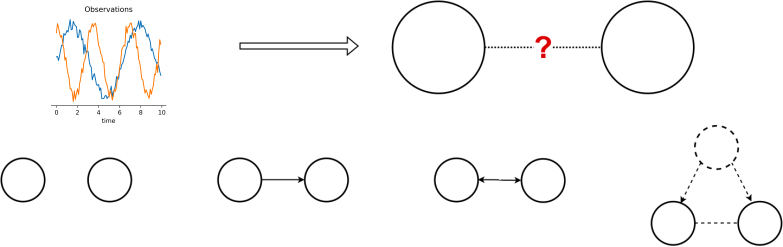}
    \caption{\textbf{Causal discovery from time series.} Causal discovery algorithms try to identify causal structures between time series, whether it is independence, unidirectional direct or circular causation, or hidden confounding.}
    \label{fig:discovery}
\end{figure}

A causal understanding of the world is crucial to building robust and explainable models for prediction and control.
Where interventions are not possible, observational causal discovery methods help to determine the underlying causal structures from static or time-series data. 
Despite its apparent simplicity, this is a non-trivial task because there can be multiple causal structures between the variables, including unidirectional causation,  bidirectional causality, hidden confounders, or even independence \cite{Pearl2000, BENKO2024115142, Runge_2018, benko_causal_2019, Stippinger2023markov, STIPPINGER2023114054}.
Further complicating the matter is the fact that the underlying systems can exhibit either stochastic or deterministic characteristics.
The task of a causal discovery algorithm is to identify the actual structure from observational data, if possible (Fig.\,\ref{fig:discovery}).

For time series, there are many approaches to discovering causal structures, such as the methods based on the predictive causality principle or the theory of nonlinear dynamical systems. Statistical and information-theoretic methods work by following the principle of predictive causality \cite{Wiener56, granger_1969, schreiber_2000, wibral_2013, geweke_1982, bressler_2011, tian_2021, Runge_2018, Runge_2019, Tank_2021, Marcinkevics_2021}.
According to this, given two time series, one is a cause of the other if its values help in predicting the future values of the other.
Granger Causality (GC) \cite{granger_1969} and Transfer Entropy (TE) \cite{schreiber_2000,wibral_2013} are two members of this family with frequency-domain extensions such as Spectral Granger Causality \cite{geweke_1982,bressler_2011} and Transfer Entropy Spectrum \cite{tian_2021}.
These methods work well for linear systems and for unidirectional couplings, but despite recent extensions \cite{Runge_2018, Runge_2019,Tank_2021, Marcinkevics_2021}, nonlinearity and hidden confounders may pose a significant obstacle to the detection of causal relationships.

The nonlinear dynamics approach leverages the properties of dynamical systems to reveal causal connections \cite{sugihara_2012, ye_2015, benko_causal_2019, hirata_2010, benko_complete_2020, benko_manifold_2020}.
Mathematical theorems \cite{takens_1981, stark_1999, deyle_2011} enable the reconstruction of system states from time series, allowing for the comparison of trajectories to discern causal relationships among variables.
Prominent methodologies within this domain include Convergent Cross-Mapping (CCM) \cite{sugihara_2012, ye_2015, benko_causal_2019}, the technique of recurrence plots \cite{hirata_2010}, and causal inference based on manifold dimensions \cite{benko_complete_2020, BENKO2024115142} or complexity measures \cite{STIPPINGER2023114054}, all of which are predominantly applied within the time domain. 
These methodologies are efficient in identifying weak directed and circular couplings, as well as revealing latent confounding factors within nonlinear systems, although their applicability is restricted to time-domain information.
The extension to the spectral domain introduces a novel dimension for causal discovery. 
In this study, we pursue this objective by extending CCM to the frequency domain.

\subsection{Dynamical systems and Convergent Cross Mapping}

CCM leverages the mathematical framework of nonlinear dynamical systems to identify causal relationships from time series data. In this section, we review the fundamental concepts of non-linear dynamical systems and the methodology of causality detection using the CCM technique.

A dynamical system constitutes a self-mapping of the state space. The subsequent state is determined by the current state in conjunction with the mapping. This mapping may exhibit deterministic or stochastic characteristics, the temporal parameter may be continuous or discrete, and the system's dynamics can be either linear or nonlinear. 
For discrete time dynamical systems, the next state is generated by the update rule:
\begin{equation}
    s_{t+1} = f(s_t)    
\end{equation}
where $f: \mathcal{S} \rightarrow \mathcal{S}$ is the mapping.
Illustrative instances of nonlinear discrete-time dynamical systems encompass the logistic map, the tent map \cite{May1976}, and the Labos neuron model \cite{Labos1994}, among others.

In the continuous-time case we can describe the dynamics by differential equations:
\begin{equation}
\dot{s} = \varphi(s)    
\end{equation}
where the time derivative of the state $s$ defines a flow in the state space. 
The trajectories of the system can be obtained from an initial condition $s_0$ by solving the differential equations $s_T = \int_0^T \varphi(s_t) \mathrm{d}t$.
Prominent examples of nonlinear continuous-time dynamical models include the Lorenz system \cite{lorenz1963deterministic}, the Rössler system \cite{Rossler1976}, the Lotka--Volterra equations \cite{Lotka1925, Volterra1928}, and the Wilson--Cowan model of cortical dynamics \cite{Cowan2016}.

In practical scenarios, direct observation of a system's state is often unfeasible, with only time-series data being accessible.
This one-dimensional data shows only a, possibly non-linear, projection of the state, and the reconstruction of the underlying high-dimensional state from these observations is thus a pivotal task. Fortunately, Takens' theorem offers a robust mathematical framework for state reconstruction from time series data \cite{takens_1981}.
Time Delay Embedding constitutes a methodology for non-linear state space reconstruction, wherein one augments a scalar time series $x(t) \in \mathbb{R}$ to reconstruct the state $s(t) \in \mathbb{R}^d$ of the underlying dynamical system:
\begin{equation}
    X(t) = [x(t), x(t-\tau), ..., x(t-(E-1)\tau)]    
\label{eq:embedding}
\end{equation}
where $X(t) \in \mathbb{R}^E$ denotes the reconstructed state, with $E \in \mathbb{N}$ and $\tau \in \mathbb{R}$ representing the embedding dimension and the embedding delay respectively. Provided that $E>2d$, the reconstruction occurs with a high probability, where $d$ signifies the intrinsic dimension of the attractor dynamics.

In the context of two interacting systems, where one system exerts influence over the other, the driver encodes its state into the state of the driven system in accordance with the skew product theorem \cite{stark_1999}.
The state of the driven system encapsulates comprehensive information about the state of the driver, whereas the converse does not hold, thereby giving the chance to identify directed causal links.

\begin{figure}[tb!]
    \centering
    \includegraphics[width=\linewidth]{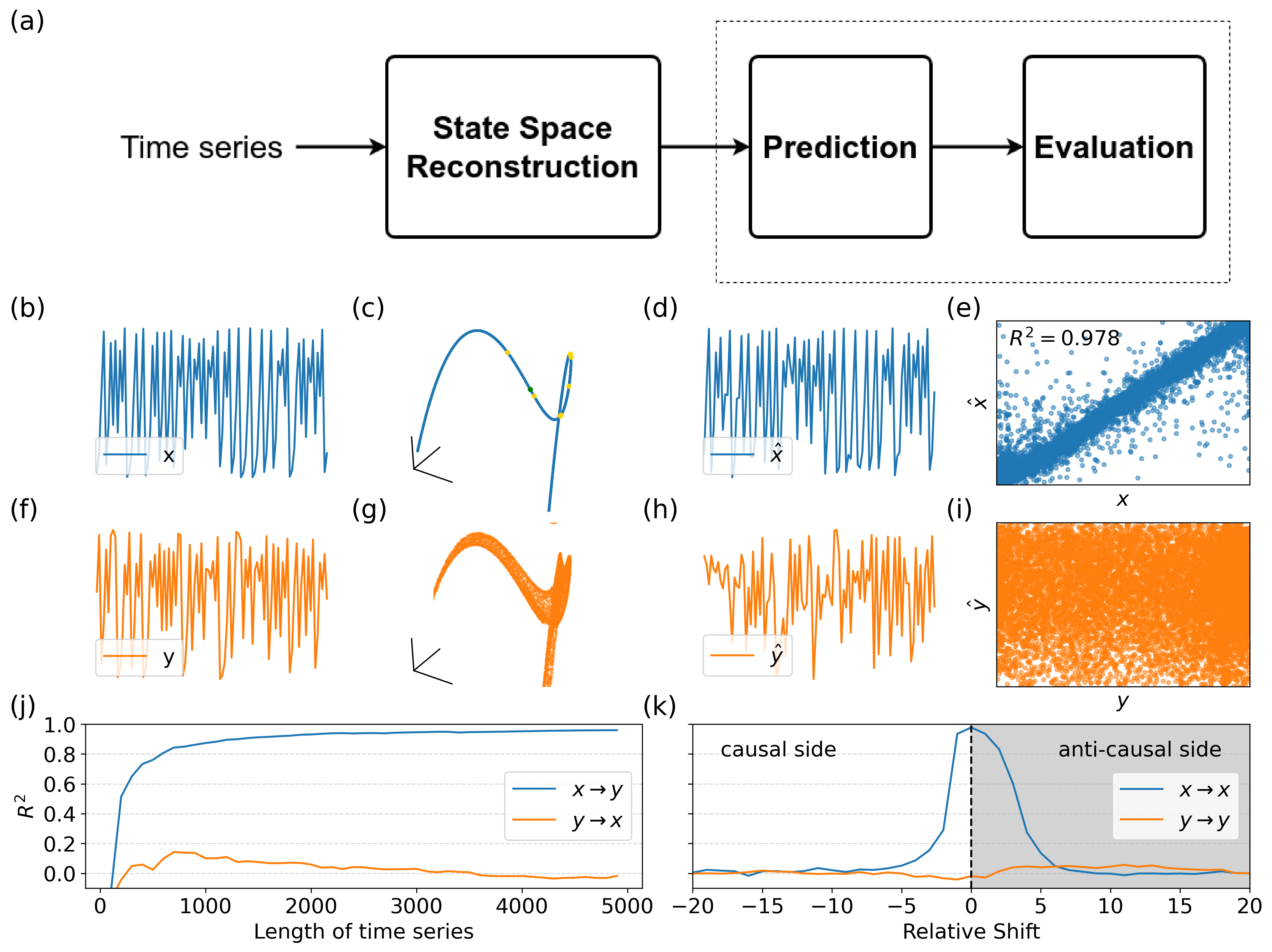}
    \caption{\textbf{Overview of Convergent Cross Mapping.}
    (a) Overview of the CCM algorithm from a macroscopic perspective and (b - k) demonstrative stages for the coupled logistic map system $x \rightarrow y$.
    (b, f) The inputs consist of individual time series observations for the $x$ (blue) and $y$ (orange) variables.
    (c, g) State-space reconstruction employing time-delay embedding. The three-dimensional plots depict the reconstructed states of the subsystems $X$ (blue) and $Y$ (orange).
    (d, h) Predicted time series $\hat{x}$ (blue) and $\hat{y}$ (orange).
    (e, i) Evaluation of the the predictions using the coefficient of determination metrics ($R^2$). The models exhibited high prediction-quality for the $x$ (blue) variable, whereas low prediction-quality for $y$ (orange) variable.
    (j) Convergence as the function of library length. The $R^2$ values are converging in the $x \rightarrow y$ direction (blue), but do not converge in the $y \rightarrow x$ direction (orange), thereby indicating a unidirectional causal link.
    (k) Cross-mapping quality as the function of time shift between the time series (Cross-Mapping Function). The negative semi-axis denotes the causal side, where the cause precedes the effect. The positive semi-axis represents the anti-causal side, where the predictability of the effect from the cause can be identified. The peak on the blue curve at zero signifies unidirectional coupling from $x$ to $y$ with minimal propagation delay. The absence of peaks in the reversed direction indicates no causal link in the $y$ to $x$ direction (orange curve).
    }
    \label{fig:CCM}
\end{figure}

Convergent Cross-Mapping (CCM) is a causal discovery algorithm grounded in Takens' and skew product theorems \cite{sugihara_2012}.
CCM deduces directed causal relationships between systems based on two time series (Box 1). 
The algorithm comprises three fundamental stages: state-space reconstruction, prediction, and evaluation (Fig.\,\ref{fig:CCM}). 
The algorithm's input consists of two time series, with the objective of investigating the interconnection between them (Fig.\,\ref{fig:CCM}a, b, f).
In the reconstruction phase, CCM employs time-delay embedding on the time series to reconstruct the state of the underlying dynamics (Fig.\,\ref{fig:CCM}a, c, g). Consequently, the manifold of each system is reconstructed. The embedding dimension ($E$) and embedding delay ($\tau$) are critical parameters of this methodology.
In the prediction phase, CCM employs weighted k-nearest neighbors (kNN) regression models to predict time series from the reconstructed manifolds in both directions (Fig.\,\ref{fig:CCM}d, h).
The inputs to the regression model consist of data points within one of the reconstructed state-spaces, while the outputs are the estimates for the corresponding time series.
The number of neighbors ($k$) is an integer parameter, determined as $E+1$.
Subsequently, the evaluation phase scrutinizes the performance of the prediction models to infer the causal relationships between the underlying systems (Fig.\,\ref{fig:CCM}e, i). The coefficient of determination $R^2$ (the squared Pearson correlation) serves as the standard metric for evaluation.
The prediction and evaluation steps are iterated across varying sample sizes to ascertain whether the coefficient of determination converges with increasing sample size (Fig.\,\ref{fig:CCM}j). Should $R^2$ converge to a significantly positive value, the predicted variable may be inferred as a driver of the model input variable. Additionally, temporal shifting of the time series and subsequent reanalysis can elucidate the propagation delay of the effect (Fig.\,\ref{fig:CCM}k).
\break

\noindent
\fbox{ \begin{minipage}{\textwidth}
\subsection*{Box 1 : Computing the CCM Score}\label{sec:ccm}
    \begin{enumerate}
        \setcounter{enumi}{-1}
        \item Time series $x_t, y_t$
        \item Time Delay Embedding ($X(t), Y(t)$)
        \item Neighborhood search $k$NN around each point in X, store neighbor indices ($\tau_j^t$) and distances ($d_{tj}$)
        \item Compute weights: $w_{tj}= \gamma_{tj}/\sum_{j=1}^k \gamma_{tj}$ , $\gamma_{tj} = \exp\left(-\frac{d_{tj}}{d_{t1}}\right)$
        \item  Mapping with computed weights: $\hat{y}_t =\sum_{j=1}^{k} w_{tj} y\left(\tau_j^t\right)$, where $\tau_j^t$ is the time index of the $j$th nearest neighbor of $X(t)$ in the reconstructed state-space.
        \item Evaluation of predicted values ($R^2= \mathrm{Corr}(\hat{y}, y)^2$)
        \item repeat $2 \rightarrow 5$ with higher library lengths (sample size), and check whether $R^2$ values are converging or not.
    \end{enumerate}

If $R^2$ converges, then $Y \rightarrow X$.
Swap the variables and repeat the computation to check in the other direction for a causal effect.
These steps fit into the basic structure of state-space reconstruction ($1.$), prediction ($2-4.$), and evaluation ($5-6.$) scheme shown in Fig.\,\ref{fig:CCM} and Fig.\,\ref{fig:CMC}. 
\end{minipage} }
\break

Over the past decade, numerous advancements have been made to the foundational CCM algorithm to augment its capabilities and enhance its robustness. The fundamental concepts of mutual cross-mapping \cite{schiff_detecting_1996} are effectively implemented by the CCM algorithm \cite{sugihara_2012} and have been subsequently refined for asymmetric inference \cite{mccracken_convergent_2014}. Additional extensions have been developed to address temporal variations in interactions \cite{huanfei_ma_detecting_2015, ge_dynamic_2021}, short parallel time series \cite{adam_thomas_clark_spatial_2015}, and potential delays in causal effects \cite{ye_2015, huang_systematic_2020}.
The delay in causal effect is ascertained by identifying the peak performance of the CCM function with respect to temporal shifts, wherein the two time series are temporally displaced relative to one another (Fig.\,\ref{fig:CCM}k, Fig.\,\ref{fig:CMC}b). 
If the optimal shift resides on the causal side of the time-shift axis, indicating that the cause precedes the effect, this signifies a delayed causal effect.
Conversely, if the highest peak of the CCM function is located on the anticausal side, it denotes a Granger causality-like predictability relationship between the variables \cite{benko_causal_2019, deng_causalized_2023}.
In the presence of predictability, the automatic determination of directed causal links becomes intricate with the current CCM methodology, necessitating measures to disentangle causal peaks from the predictability effects.

We delineate additional advancements within the reconstruction-prediction-evaluation framework. To enhance state space reconstruction in the CCM paradigm, methodologies such as Gaussian processes \cite{feng_improving_2020, feng_detecting_2019}, reservoir computing \cite{huang_detecting_2020}, recurrent neural networks \cite{brouwer_latent_2021}, and frequency domain embedding \cite{avvaru_2023} have been proposed.

To enhance the evaluation and prediction phases, efforts were made to automate the evaluation procedure and refine the evaluation metric. To eliminate the subjective element in determining the convergence, automatic fitting of convergence rates was implemented \cite{monster_causal_2017}. Multiple attempts were made to develop a significance test for the framework, although with limited success \cite{wang_improved_2019, krishna_inferring_2019, huang_systematic_2020, sarah_cobey_limits_2016, krakovska_testing_2016}. Alternative prediction methodologies and evaluation metrics have been proposed as modifications to the foundational algorithm to enhance robustness, including symbolic CCM \cite{ge_symbolic_2023}, Convergent Cross Sorting \cite{breston_convergent_2021}, Gaussian Process priors \cite{ghouse_inferring_2021, ghouse_inferring_2022}, Neural Shadow Mapping \cite{vowels_shadow-mapping_2021}, Robust CCM \cite{diaz_inferring_2022}, and a metalevel approach via the Continuity scaling method \cite{ying_xiong_continuity_2022}.

These extensions predominantly addressed the bivariate context, and alternative methodologies were developed to generalize CCM to multivariate signals, including partial cross-mapping \cite{leng_partial_2020} and other techniques \cite{van_berkel_modeling_2021, nithya_multivariable_2021, krishna_inferring_2019}.

Despite substantial advancements in the Convergent Cross-Mapping (CCM) algorithm, progress in the domain of spectral causal discovery remains limited.
To date, only two methodologies, Frequency-Domain CCM (FDCCM)\cite{avvaru_2023} and Cross-Frequency Symbolic CCM (CFSCCM)\cite{gu_detection_2023}, have made strides in this area.
Frequency-Domain Convergent Cross-Mapping utilizes a frequency-domain state representation for the reconstruction of state space. However, this approach is not designed to discern frequency-specific causal interactions.
Cross-Frequency Symbolic CCM, on the other hand, employs the Hilbert transform to decompose temporal signals and evaluates CCM between the resultant signal modes, thereby examining phase-amplitude couplings.

In this study, we introduce Cross-Mapping Coherence (CMC), an extension of Convergent Cross Mapping (CCM) into the frequency domain for the identification of frequency-specific causal relationships between time series.
Additionally, we propose a novel methodology to disentangle predictability from dynamic causality.
In the following sections, we delineate Cross-Mapping Coherence and demonstrate its efficacy on coupled systems such as the Logistic map, the Lorenz system, Kuramoto oscillators, and the Wilson-Cowan model.
Furthermore, we evaluate its robustness with respect to the choice of embedding parameters and the presence of observational noise.
We contextualize CMC within the framework of spectral causal discovery methods, highlighting its merits and limitations.
Lastly, we provide a comprehensive description of simulations and analyses in the Methods section.

\section{Results}

    \subsection{Cross-Mapping Coherence}
    \begin{figure}
        \centering
        \includegraphics[width=\textwidth]{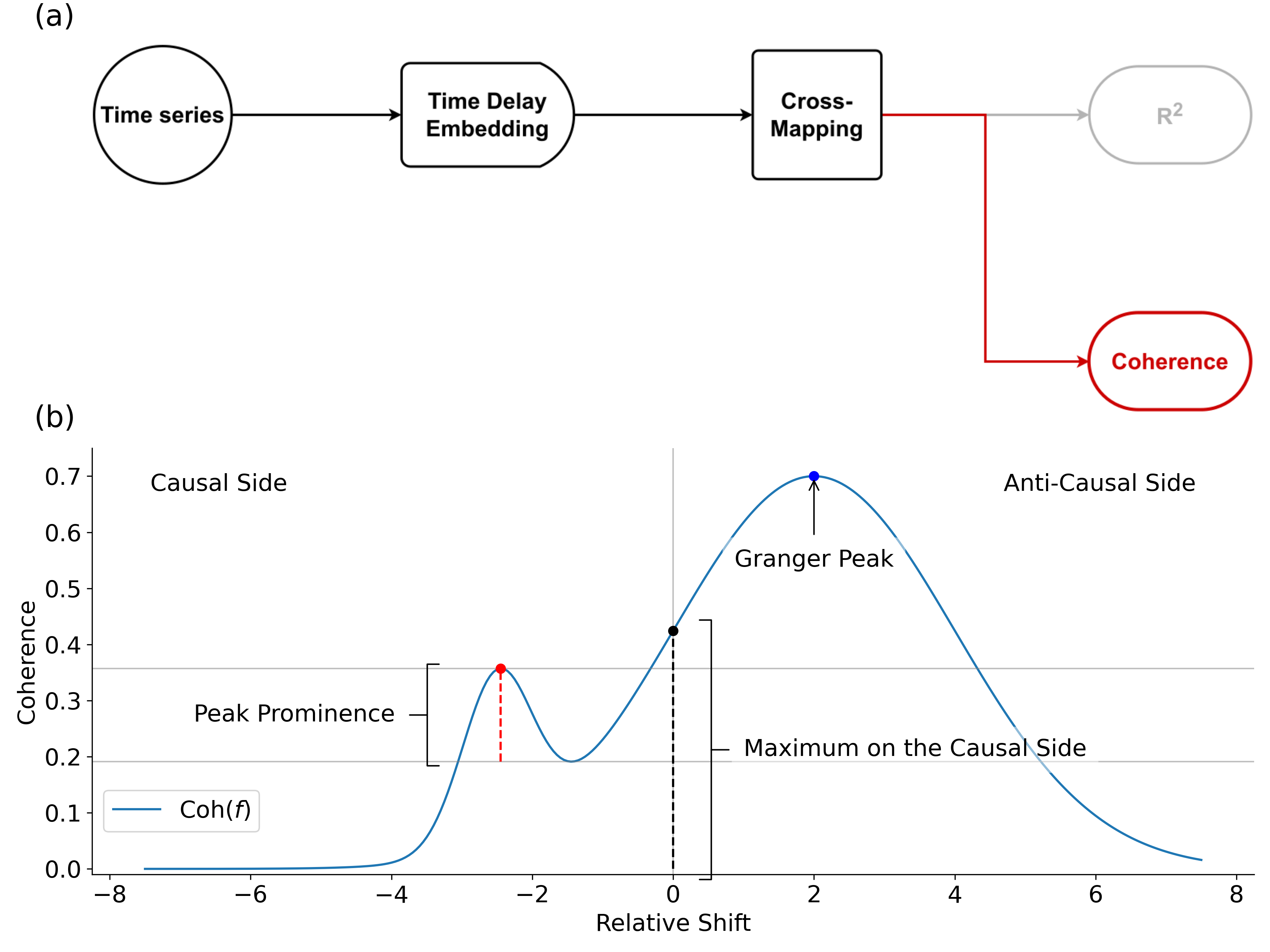}
        \caption{\textbf{Cross-Mapping Coherence detects causal relationship in the frequency domain and filters out spurious causality by peak prominence calculation.}
        (a) A comparative overview of Cross-Mapping Coherence (CMC) and Convergent Cross Mapping (CCM).
        Both methodologies analyze two time series and are comprised of three principal stages: state reconstruction, prediction, and evaluation.
        Although the state reconstruction and prediction phases are identical in both methods, CMC diverges in the evaluation phase.
        Specifically, CMC utilizes the coherence metric to compare predicted and actual values, thereby offering spectral insights into the causal connection.
        (b) Significant peak determination with peak prominence calculation on the Cross-Mapping Coherence Function. 
        The coherence value is computed for each temporal shift, followed by an examination of the resultant coherence function across individual frequency bands. An illustrative coherence function is presented at a specific frequency (blue curve). Peaks may manifest on the anti-causal side of the function (Granger peak; blue dot) if the state of the effect can be inferred from the state of the cause.
        This Granger peak may hinder the causal inference procedure, so if predictability is present, one has to handle maximum determination with special care.
        To reduce the impact of Granger peaks, we identify peaks on the causal side of the graph, and calculate the prominence value of each peak to quantify the causal effect.
        }
        \label{fig:CMC}
    \end{figure}
    
    
    


    We expand the Convergent Cross-Mapping (CCM) method to the frequency domain, allowing us to identify causal links in specific spectral bands.
    Cross-Mapping Coherence (CMC)  follows the same basic steps as CCM – reconstruction, prediction, and evaluation – but the evaluation step is altered (Fig.\,\ref{fig:CMC}a).

    To apply CCM, one takes two time series, applying time delay embedding to reconstruct the state from both (Fig.\,\ref{fig:CCM}).
    Then we employ the cross-mapping step with the k-NN method to predict each other's states.
    Finally, we evaluate the quality of the prediction by computing the coefficient of determination between the actual and predicted values.
    To determine the causal effect delay, we shift the time series relative to each other, compute the CCM values, and find the maximum of the resulting CCM function.
    
    With CMC, the reconstruction and prediction steps remain the same, but the evaluation process is different (Fig.\,\ref{fig:CMC}).
    Rather than using the coefficient of determination, we calculate the coherence between the predicted and actual time series (Eq.\,\ref{eq:coherence}), providing spectral information about the quality of the prediction:
    
   \begin{equation}\label{eq:coherence}
        \mathrm{coh}_{x\hat{x}}(f) = \frac{|S_{x\hat{x}}(f)|}{ \sqrt{ S_{xx}(f) S_{\hat{x}\hat{x}}(f) } }
    \end{equation}
    where the coherence $\mathrm{coh}_{x\hat{x}}$ at a specific frequency $f$ is defined as the magnitude of the averaged cross-spectral density between the time series $x$ and the predicted values $\hat{x}$, normalized by the square root of the power spectral densities $S_{xx}$ and $S_{\hat{x}\hat{x}}$ of the respective signals \cite{bastos_tutorial_2016}.

    To determine the causal effect delay, we shift the time series relative to each other and recompute the cross-mapping coherence metric resulting in a 2D shift- and frequency-dependent CMC function.

    \subsection{Effect Delay Determination with Peak Prominence}
    The time delay dependence of the CCM, as well as CMC, raises the problem of significant maxima detection: while we can easily assume that higher correlation and coherence values correspond to stronger connections, simply taking the maximum of the CCM or the CMC functions does not produce satisfactory results (Fig.\,\ref{fig:CMC}b). The problem is, that high values can appear not only because the cause can be reconstructed from the consequence, but because the consequence is predictable based on the cause. Consequently, high peaks can occur at positive time delays, where the consequence precedes the cause, resulting in apparent contradictions; we call these peaks Granger peaks. While it is clear that these peaks should be excluded from the evaluation, they make it more difficult to select the proper peaks, as side peaks on the causal semi-axis (where cause precedes consequence) can be formed due to random fluctuations on the shoulder of Granger peaks on the positive semi-axis. To make the task even more complex, both causal and Granger peaks can appear simultaneously, especially in the case of bidirectional coupling.
    
    To find the amplitude of the most significant peak on the negative semi-axis, even in the presence of a possible Granger peak on the positive semi-axis, the prominence of each local peak on the delay-dependent CMC functions is calculated for each frequency band.
    The prominence measure of a peak is used especially in geography and it expresses how much higher a peak is over its area of dominance.
    Specifically, the prominence of a peak corresponds to its height over the minimum, which should be crossed to get to a higher peak, while the prominence of the absolute maximum is equal to its elevation.

    We measure the strength and delay of the causal relationship, for each frequency band, by the prominence and delay of the most prominent peak of the coherence values, where it precedes or co-occurs with the consequence. This case holds if the time index of the peak is less than the embedding time window (E in Eq.\,\ref{eq:embedding}), wherein positive indices are permitted in recognition of the temporal uncertainty induced by time delay embedding.
    Thus, if the time index of the absolute maximum is less than $E$, then the momentary causal strength is expressed by its value, but if the time index of the absolute maximum is above $E$, the causal strength is less, corresponding only to the prominence of a peak below $E$.
    In this way, we only get high causality if a well-identifiable independent peak occurs below $E$.

    The prominence-based computation facilitates the identification of a genuine peak on the negative semi-axis, whilst preventing the erroneous recognition of minor local maxima situated on the shoulder of a more substantial peak on the positive semi-axis.
    
    In the following subsections, we demonstrate the effectiveness of CMC on time series of simulated systems.

    \subsection{Logistic maps}

        \begin{figure}[htb!]
            \centering
            \includegraphics[width=\textwidth]{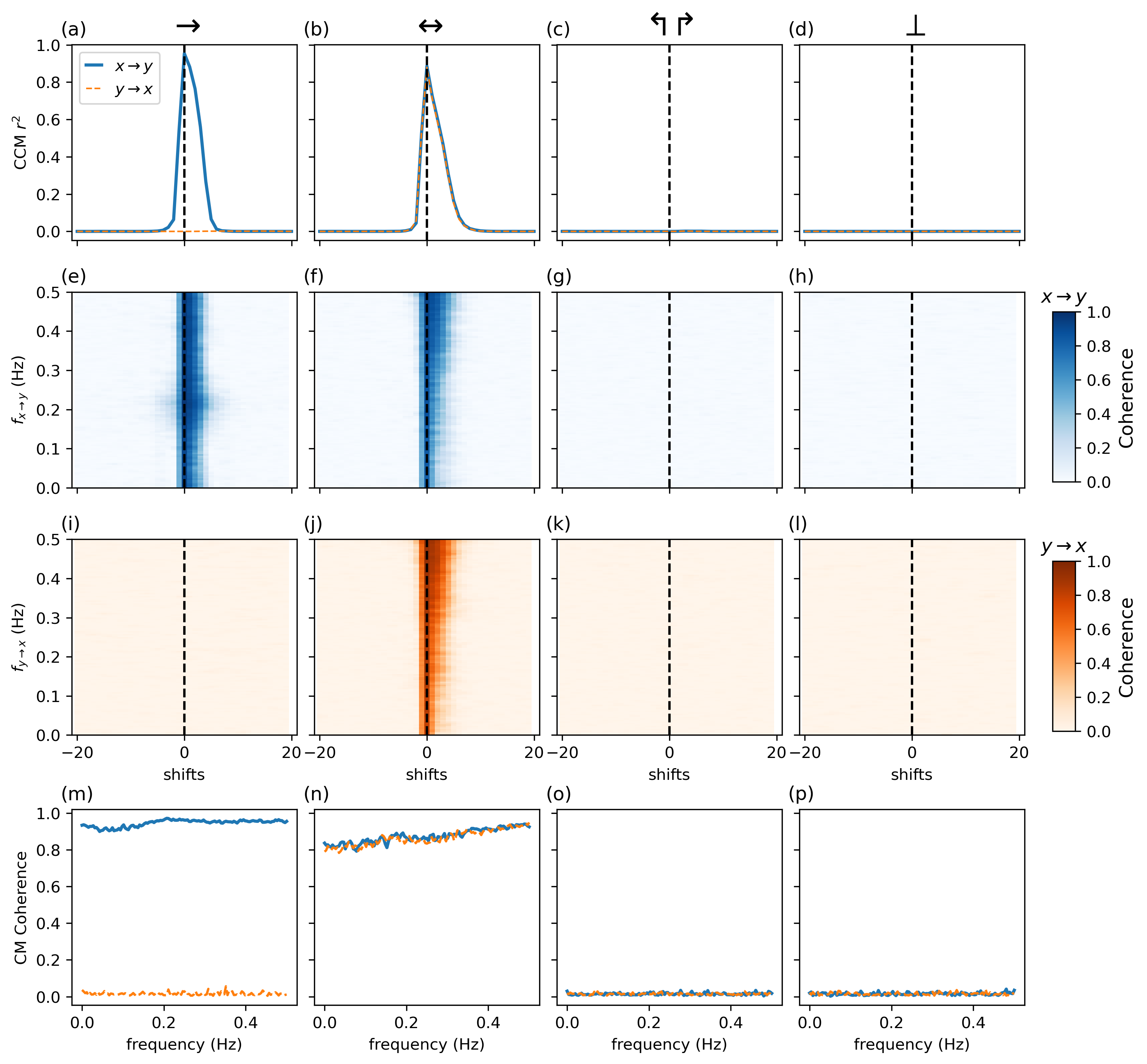}
            \caption{\textbf{Causal discovery on the logistic map systems with diverse coupling scenarios using Cross-Mapping Coherence.}
           The columns of the figure illustrate various coupling configurations: unidirectional coupling $\rightarrow$ (a, e, i, m), circular coupling $x \leftrightarrow y$ (b, f, j, n), hidden common driver $x \Lsh \Rsh y$ (c, g, k, o), and independence $x \perp y$ (d, h, l, p), respectively. The direction of the inferred relationship is color-coded: the $x \rightarrow y$ direction is indicated by blue, whereas the $y \rightarrow x$ direction is marked by orange colors.
            (a-d) The time-delayed CCM function reveals causal relations from bivariate time series.
            A pronounced peak indicates the presence of causal links in both the unidirectional  (a) and circular (b) scenarios.
            However, the absence of distinct peaks in the hidden common driver (c) and independent (d) cases mark the lack of direct causal connections.
            (e-l) CMC functions for the coupled logistic map systems show spectral information for the inferred $x \rightarrow y$ (e-h) and $y \rightarrow x$ (i-l) directions. The vertical bands on the two-dimensional CMC functions in the case of unidirectional and circular couplings indicate a roughly frequency-independent causal link between the time series (e, f, j). The absence of such a band indicates that there were no direct causal link to be inferred (i). Here the colormaps are normalized to the $[0, 1]$ interval.
            (m-p) The peak prominence of CMC values show the proper causal couplings uniformly along the frequency axis.
            }
            \label{fig:logmap}
        \end{figure}

        To test the CMC method, we performed CCM and CMC analyses on bivariate logistic map systems with unidirectional, circular, hidden driver, and independent couplings.

        The time-delayed CCM analysis recovered the correct direct causal connections from the time series, and the CMC investigation revealed near-uniform spectral contributions (Fig.\,\ref{fig:logmap}).
        In the unidirectional and circular coupling cases, we found high coherence values at every frequency band, while we observed near-zero coherence values for the hidden driver and independent case.

        \begin{figure}
            \centering
            \includegraphics[width=\linewidth]{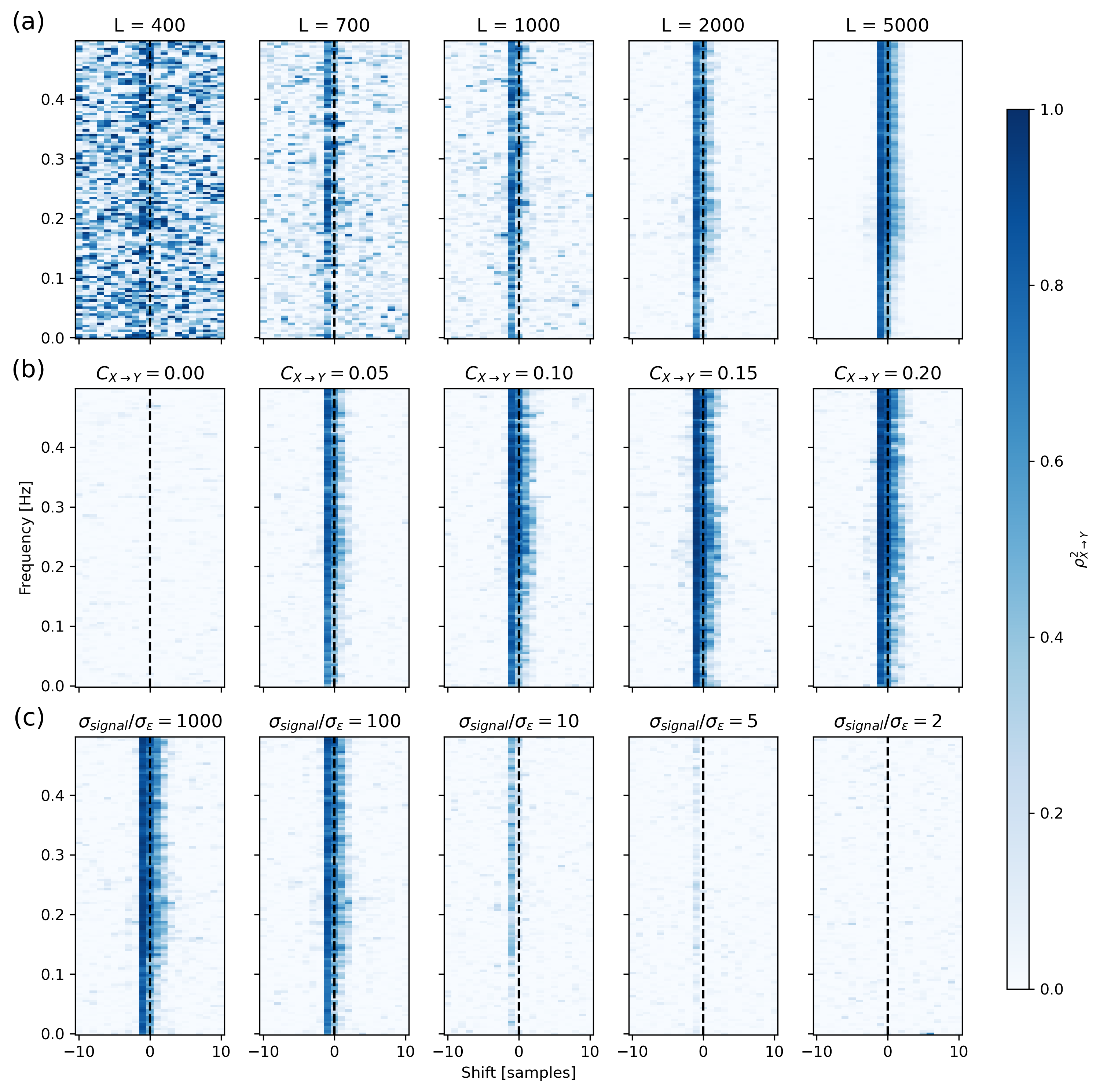}
            \caption{\textbf{Dependence of CMC on simulation parameters for logistic maps with unidirectional coupling.}
            (a) Dependence on time-series length ($D=2$, $C_{X \rightarrow Y}=0.15$, $\sigma_{\varepsilon}=0$). The CMC function is convergent, the causal link is hardly detectable for $L=400$, but clearly visible at $L=700$ simulation steps.
            (b) Dependence on coupling strength ($L=2000$, $D=2$, $\sigma_{\varepsilon}=0$). The causal link is detected by CMC even in case of such weak coupling as $C_{X \rightarrow Y}=0.05$.  (c) Dependence on signal-to-noise ratio ($L=10000$, $D=2$, $C_{X \rightarrow Y}=0.15$ ). CMC can detect the causal link when the signal-to-noise ratio is at least between $5$ and $10$.}
            \label{fig:length_coupling_noise}
        \end{figure}

        \begin{figure}
            \centering
            \includegraphics[width=\linewidth]{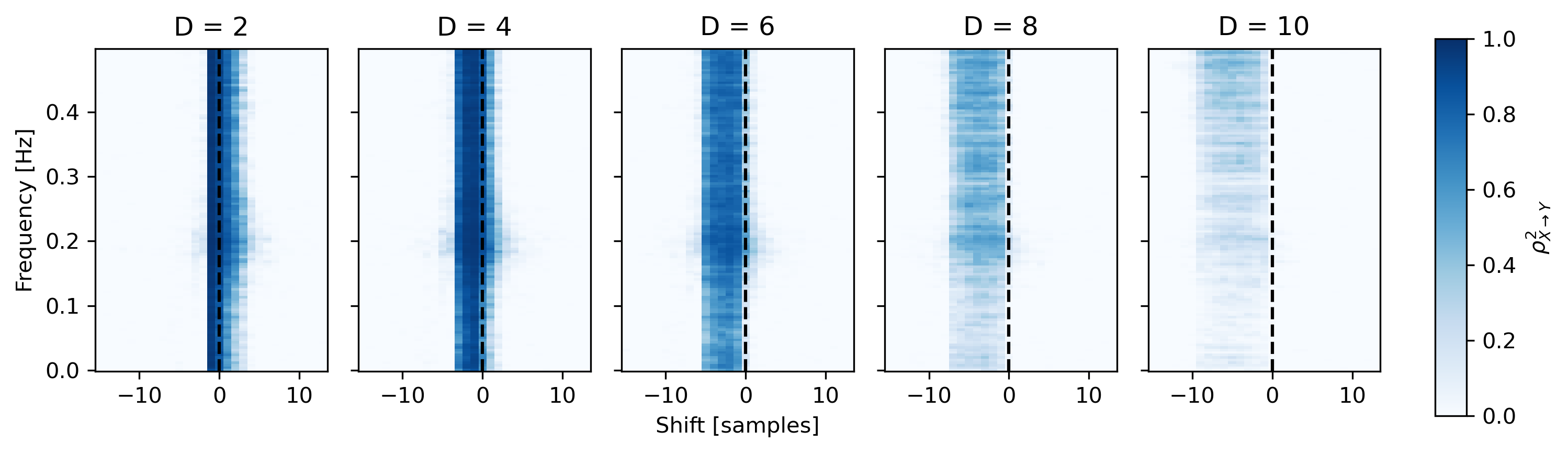}
            \caption{\textbf{Dependence of the CMC function on the embedding dimension.}
            Time delay embedding introduces temporal uncertainty in the determination of the effect-propagation delay, thus higher embedding dimensions result in a plateau in the CMC function proportional to the embedding window size ($w=(D-1)\tau$).}
            \label{fig:embedding_dim}
        \end{figure}

        We investigated the dependence of CMC on simulation parameters for the unidirectional coupling case (Fig.\,\ref{fig:length_coupling_noise}).
        First, we looked at CMC as a function of time series length by computing the performance on various sample sizes from $L=400$ to $L=5000$ with $E=2$ and $\tau=1$ embedding parameters.
        CMC was convergent: the directional coupling is not detectable on low sample size ($L=400$), but the pattern of connection becomes clear from a sample size of $L=1000$ (Fig.\,\ref{fig:length_coupling_noise}a).
        Second, to see how cross-mapping performance depends on the coupling strength between the systems, we simulated unidirectionally coupled logistic map systems ($L=2000$, $E=2$, $\tau=1$) by varying the value of the coupling parameter $C_{X \rightarrow Y}$ in the range $[0, 0.2]$. 
        CMC proved to be very sensitive, as it could detect even weak coupling strengths ($C_{X \rightarrow Y}=0.05$) between the subsystems (Fig.\,\ref{fig:length_coupling_noise}b).
        Third, the noise robustness of the CMC method was tested by adding observational noise to the time series ($L=2000$, $E=2$, $\tau=1$), varying the signal to noise ratio from $1000$ to $2$.
        CMC robustly detected the coupling down to the signal to noise value of $10$ (Fig.\,\ref{fig:length_coupling_noise}c).

        We also examined the dependence of CMC performance on the embedding parameters for the unidirectionally coupled logistic map system (Fig.\,\ref{fig:embedding_dim}).
        Varying the embedding dimension between $D=2 - 10$ ($L=10000$, $\tau=1$), we found that as the dimension increased, a plateau formed on the CMC function along the time-shift axis proportional to the size of the embedding window, while the amplitude of the coherence decreased.

    \subsection{Lorenz system}
        \begin{figure}[htb!]
            \centering
            \includegraphics[width=\textwidth]{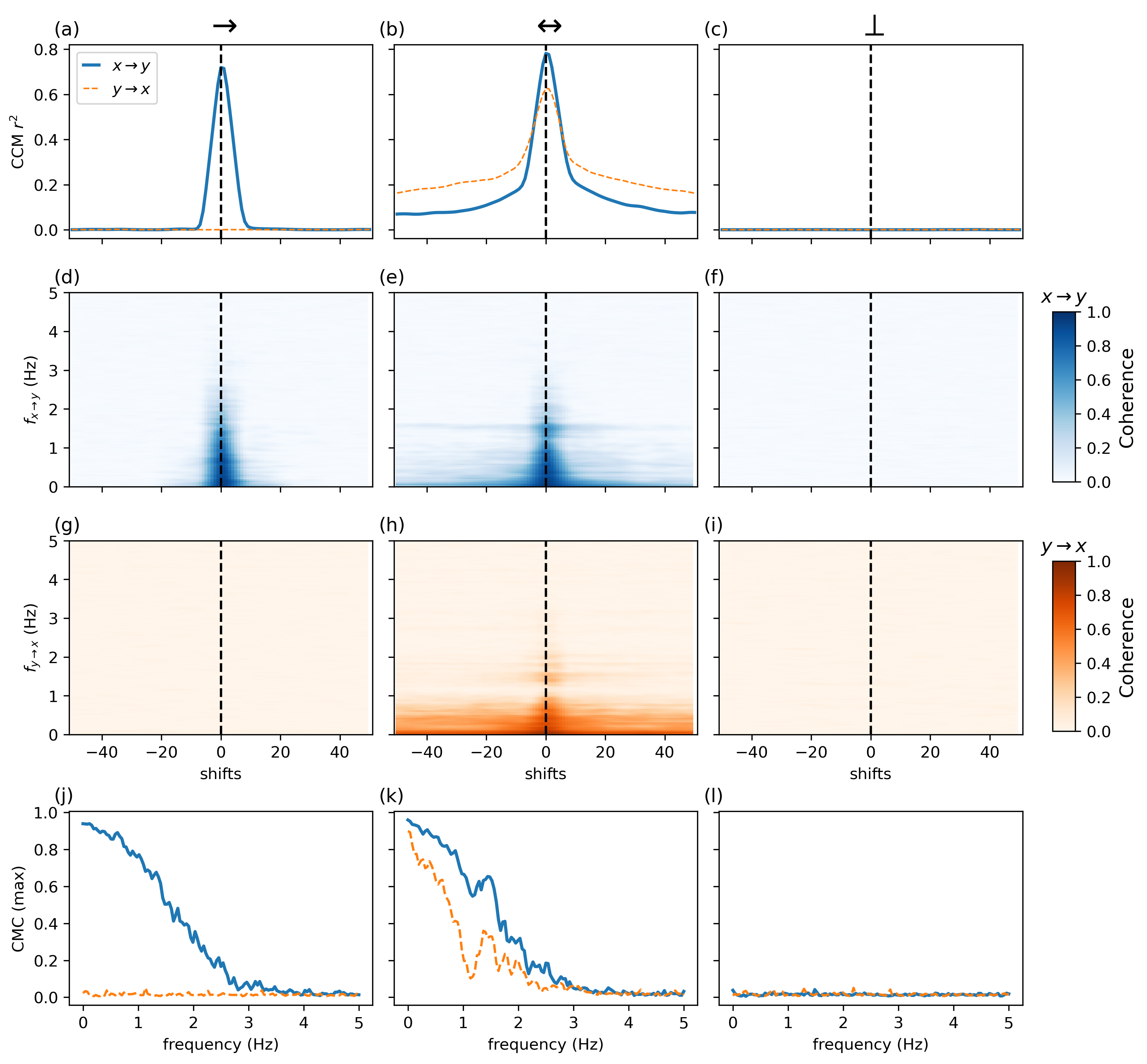}
            \caption{\textbf{Causal discovery of the Lorenz system in various coupling scenarios.}
            The columns contain different couplings like unidirectional link $\rightarrow$ (a, d, g, j), circular coupling $\leftrightarrow$ (b, e, h, k),  and independence $\perp$ (c, f, j, l), respectively.
            (a-c) The time-delayed CCM score reveals causal relations from bivariate time series.
            A peak shows the presence of causal links in the unidirectional (a) and circular (b) cases, whereas the absence of the peak marks the lack of connection in the independent case (c).
            (d-i) CMC function hints the absence of causal links. CMC scores for the $x_1 \rightarrow x_2$ direction (d-f) are drawn as blue (2nd row) and the $x_2 \rightarrow x_1$ direction (g-i) as orange (3rd row).
            (j-l) The peak prominence CMC values show the true causal couplings along the frequency axis.}
            \label{fig:lorenz}
        \end{figure}

        We carried out the CCM and CMC analysis on coupled Lorenz systems to test CMC on continuous-time dynamics (Fig.\,\ref{fig:lorenz}).
        
        The time-delayed CCM correctly detected the couplings, and CMC returned the frequency-dependent interactions.
        In the unidirectional case, a peak can be identified at $0$ timeshift for the $x \rightarrow y$ direction, and the power decreases from lower to higher frequencies.
        In the case of circular coupling, back-and-forth connections are detected, featuring a decreasing coherence spectrum with a small peak between $1$ and $2$ Hz.
        When the subsystems were independent, the CCM and CMC values were close to zero.

        \begin{figure}[htb!]
            \centering
            \includegraphics[width=\linewidth]{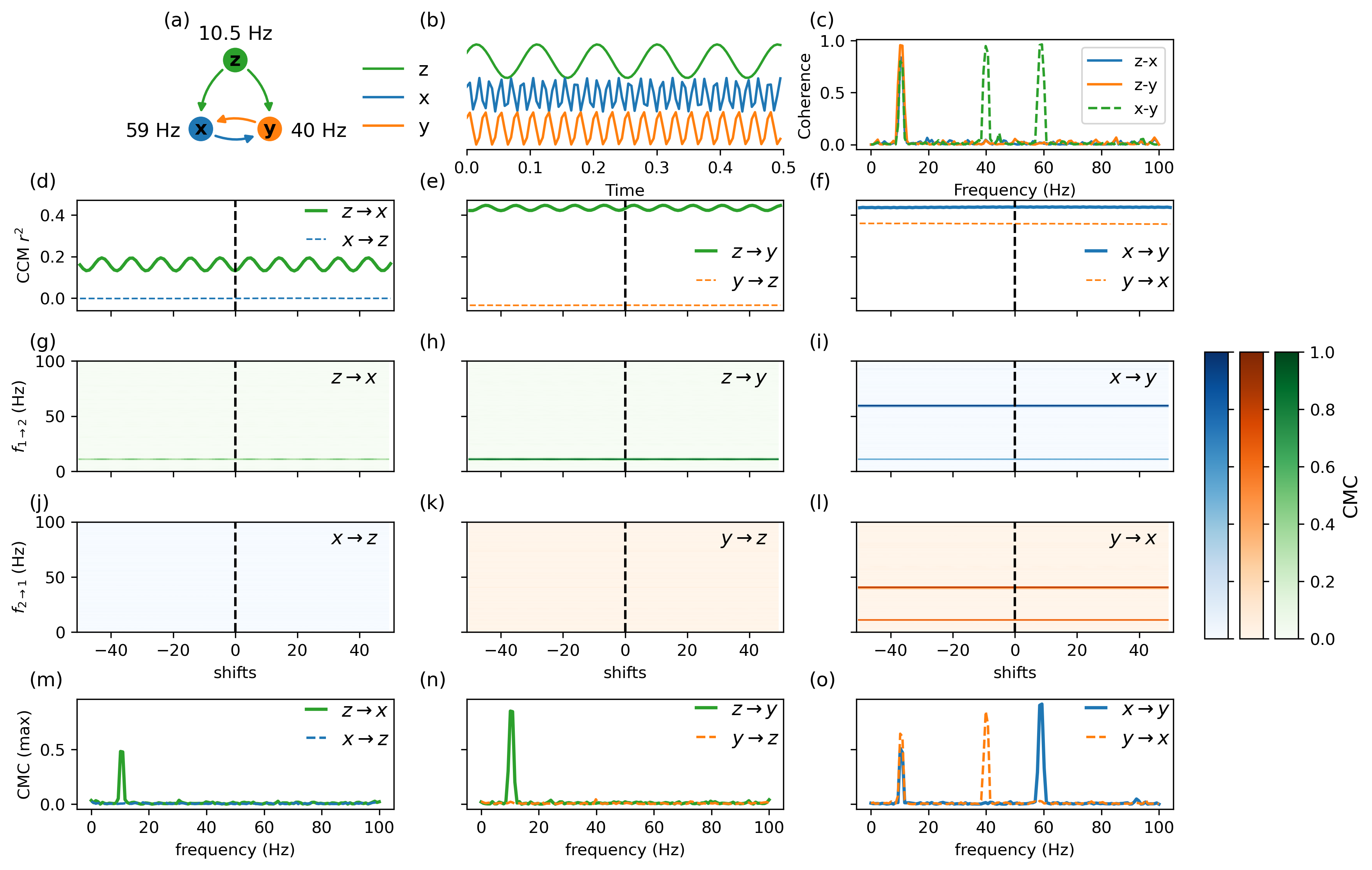}
            \caption{\textbf{CMC causal discovery on a network of Kuramoto oscillators.}
            (a) Network structure of coupled Kuramoto oscillators; the $z$ variable drives the mutually coupled $x$ and $y$ variables.
            (b) Illustrative time series from the three Kuramoto oscillators showing sinusoid waveforms of the $x$ (blue), $y$ (orange) and $z$ (green) oscillators with differing central frequencies of $59$ Hz, $40$ Hz and around $10.5$ Hz, respectively.
            (c) Coherence between the signals returns the coupling between the subsystems for the signal pairs. 
            The coherence has a single peak in for the $z-x$ (blue) and $z-y$ (orange) pairs around $10$ Hz, whereas for the $x-z$ pair the coherence function has three peaks around the central frequencies.
            (d - o) The three columns present the causality analysis results for the $x - z$ (d, g, j, m), the $y - z$ (e, h, k, n) and the $x - y$ (f, i, l, o) pairs.
            (d - f) CCM identifies directional couplings between the subsystems. 
            Specifically, it shows unidirectional coupling from $z$ to (d) $x$ and (e) $y$, and bidirectional coupling for the $x-y$ pair (f).
            (g - l) CMC scores reveal the frequency-specificity of causal links. 
            Causal effect of $z$ on $x$ and $y$ shows up as increased CMC around $10$ Hz (g, h), whereas the lack of feedback is indicated by near-zero CMC values in the backward direction (j, k).
            (i, l) CMC detects circular causality for the $x-y$ pair on distinct frequency channels. In the $x \rightarrow y$ direction the CMC function indicates causal coupling at $59$ Hz and around $10$ Hz (i). Meanwhile in the reverse $y \rightarrow x$ direction a backward causal link appears at $40$ Hz and around $10$ Hz (l).
            (m-o) Reducing the CMC function to the maximal peak, CMC values per frequency bands reveal the actual causal couplings along the frequency axis. }
            \label{fig:kuramoto}
        \end{figure}

    \subsection{Kuramoto models}

        We demonstrate how CMC revealed frequency-specific interactions on a small network of $3$ interconnected Kuramoto oscillators.
        The network was designed such that there is a common driver ($z$) with a central frequency of $10.50422624$ Hz and there are two other oscillators, $x$ and $y$ with respective central frequencies of $59$ and $40$ Hz, that also drive each other.
        
        For the $z$-$x$ and $z$-$y$ pairs, CMC showed maximum coherence at around $10$ Hz with $z$ as the driver (Fig.\,\ref{fig:kuramoto}).
        For the $x$-$y$ pair the $x \rightarrow y$ link had a peak at $60$ Hz and $10$ Hz, whereas the opposite direction had a peak at $40$ Hz and $10$ Hz.
        
        We found that CMC could recover each causal connection properly in a frequency-specific way, even the circular coupling in different frequency bands, but it could not recover the proper effect delay with certainty. 
        Due to the strong periodicity of these oscillators, both CCM and CMC had multiple peaks of similar amplitude that were periodic in the time shifts between the two time series. 
        Thus, the proper time delay of the causal effects could not be inferred. However, this did not affect the inference of the directionality and the frequency dependency of the causal effects.

    \subsection{Wilson-Cowan model}
    
        \begin{figure}[htb!]
            \centering
            \includegraphics[width=\textwidth]{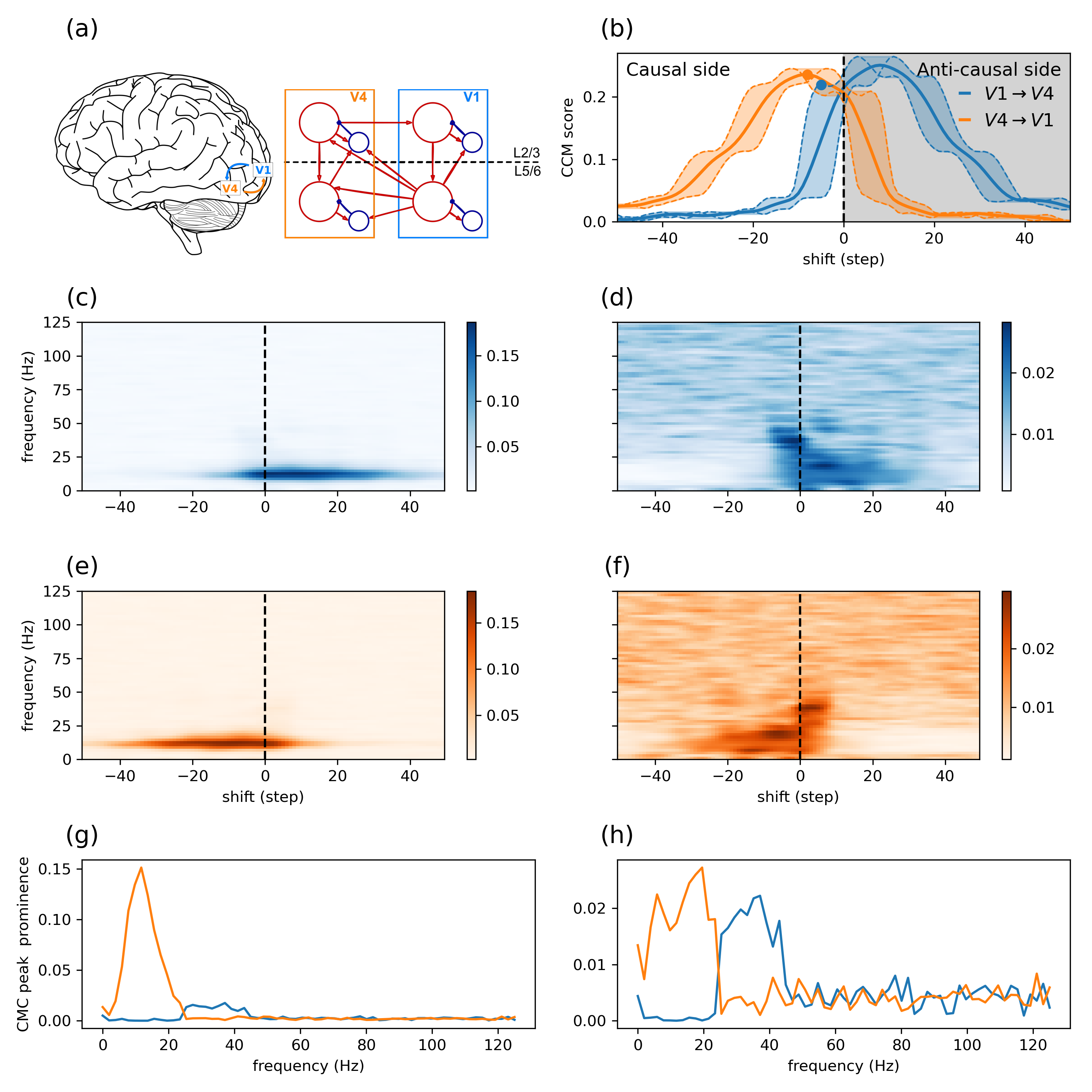}
            \caption{ \textbf{Causal discovery with CCM and CMC on a Wilson-Cowan model of cortical areas V1 and V4.}
            (a) Schematic illustration of the network. Each area is modeled by $2\times2$ stochastic differential equations. The coupling strengths are chosen to be biologically appropriate.
            (b) Time-delayed CCM analysis of the Wilson-Cowen dataset ($E=9, \tau=1$). The solid curves show the average performance over reconstructed state coordinates and realizations ($N=10$) in the function of temporal shift, whereas dashed lines denotes the score for the first and the last coordinates of the embeddings, with the transition between these latter two marked as shading giving a sense about the temporal uncertainty of effect propagation delay.
            The $V4 \rightarrow V1$ connection is stronger at the $-8$ timesteps shift (orange).
            (c, e) Shift-frequency CMC analysis shows a strong peak at the $5-20$ Hz range for the $V4 \rightarrow V1$ connection on (e) and the corresponding Granger-peak on the anti-causal side on (c).
            (d, f) The same peak is present on the band-normalized CMC functions on (f), but an additional peak becomes visible in the opposite direction ($V1 \rightarrow V4$) at the $20-50$ Hz range on (d).
            (g, h) CMC prominences reveal frequency-specific causal interactions between V1 and V4 both for the raw (g) and the normalized (h) case. From V1 to V4, the causal effect appears in the $20-50$ Hz range, whereas the V4 to V1 coupling takes shape in the $1-20$ Hz frequency band. 
            }
            \label{fig:wc}
        \end{figure}
        We applied CMC on the Wilson-Cowan model of a pair of cortical areas, V1 and V4.
        The model is a system of stochastic differential equations, that simulates mean-field cortical dynamics as coupled supragranular and infragranular excitatory and inhibitory cell populations, $2\times2$ in each area. The areas are linked together with a biologically realistic connection structure (Fig.\,\ref{fig:wc}a, \cite{mejias2016}). 
        
        CMC revealed frequency-dependent interactions in the V1-V4 model (Fig.\,\ref{fig:wc}).
        The basic time-delayed CCM analysis returned a pronounced effect from V4 to V1, without information on the frequency of the connections (Fig.\,\ref{fig:wc}b). 
        The time delay of the effect is around $-6$ time-steps, that is at $32$ ms.
        CMC found frequency-dependent connections, a feedforward V1$\rightarrow$V4 link in the $25-50$ Hz band, and a feedback V4$\rightarrow$V1 link in the $1-20$ Hz band (Fig.\,\ref{fig:wc} c-h).
        This directionality in the coupling showed up if we took the maximal prominence value of coherence over the time-shift axis(Fig.\,\ref{fig:wc}c, e, g), and became even more visible if we normalized the coherence values for each frequency band beforehand (Fig.\,\ref{fig:wc}d, f, h).
        The time shifts of the maximal prominence values were in the $-2$ and $-8$ shifts for the  V1$\rightarrow$V4 and V4$\rightarrow$V1 links respectively for the unnormalized CMC function, and at the $-1$ and $-6$ shifts for the normalized CMC analysis. 
        Both results show that the V1$\rightarrow$V4 feedforward connection has smaller effect propagation delay.

\section{Discussion}

Cross-Mapping Coherence (CMC) reveals frequency-specific interactions, but there are several ways to form a frequency-dependent method from Convergent Cross-Mapping (CCM) by making modifications at various points in the algorithm, such as the embedding, the prediction, or the evaluation steps.
Here we chose to change the evaluation metrics from correlation to coherence, because it is a minimal working modification to the original algorithm which achieves a frequency-dependent measure of causality.

We introduced the peak prominence metrics to determine causal effect delays in the frequency- and shift-dependent CMC functions.
The frequency-dependent extension of the cross-mapping method rendered the evaluation of the delay-dependent cross-mapping functions a complex task.
Thus, visual examination of the peaks for all frequency bands was not a viable solution in determining the causal effect delays in each frequency band.
To handle this challenge, we introduced a maximum detection method, based on the prominence measure of the peaks of the cross-mapping functions, applied in all frequency bands separately. Our simulations with the Wilson-Cowan model highlight the importance of this type of peak evaluation and maximum detection. 
If one would simply take the maxima of the CMC function on the negative semi-axis (where cause precedes the consequences), one would get similarly high values in both directions on all frequency bands.
If done this way, the evaluation would almost certainly return false circular results in the case of a unidirectional connection.
But closer examination reveals that, on many frequencies, the high values on the negative semi-axis are actually the upward slope of higher peaks on the positive semi-axis, thus evaluating the predictability of the consequence, not the reconstructability of the cause (Fig.\,\ref{fig:wc}c-f).  Our prominence-based solution retrieved the proper directionality for unidirectional connections, while not losing the ability to detect bidirectional connections.

CMC identifies causal relationships in the frequency domain; however, it is not unique, as other spectral methodologies, such as spectral Granger Causality (SGC) and Transfer Entropy Spectrum (TES), exhibit analogous characteristics and functionalities.
In particular, both CMC and SGC showed the same frequency-specific directional coupling in the Wilson-Cowan model \cite{mejias2016, varga_2021}.
However, given that SGC and TES are based on the predictive causality principle, whereas CMC is rooted in the theory of dynamical systems, CMC introduces a novel theoretical and practical aspect for discovering causal relationships from time-series data.

Frequency Domain CCM (FDCCM) and Cross-Frequency Symbolic CCM (CFSCCM) are two alternative extensions of the CCM method that utilize spectral information \cite{avvaru_2023, gu_detection_2023}.
FDCCM uses the Fourier transform at the embedding step, but does not provide explicit frequency-specific information about the causal link.
CFSCCM investigates the cross-frequency coupling between fast and slow signal modes.
It decomposes the signal into modes, then computes amplitudes and phases by Hilbert transform, and finally, it computes symbolic CCM between amplitudes and phases.
In contrast, CMC applies a different approach, it runs on the time domain signals and uses spectral information only at the evaluation step by the coherence metric.

CMC shows convergence with increasing data length; in the unidirectionally coupled example, this convergence seemed independent of frequency bands.
CMC may need longer measurements for systems with smoothly varying dynamics to achieve convergence, especially in lower frequency bands. To ensure the required number of recurrences, at least 15-20 cycles should occur during the investigated section to make the coherence computation meaningful.

The CMC method proved very sensitive, detecting even weak couplings. This feature is inherited from CCM since it was originally developed to detect causal relations between weakly coupled nonlinear and non-separable dynamical systems \cite{sugihara_2012}.

The CMC method demonstrates robust performance in the presence of observational noise, maintaining efficacy up to a signal-to-noise ratio of $10$.
Although this degree of noise tolerance may be adequate for specific applications, information-theoretic methods such as SGC and TES may exhibit superior performance on noisy datasets, as evidenced by recent time domain information-theoretical methods \cite{PhysRevE.102.062201}.

The choice of embedding parameters affects the shape of the CMC function.
The embedding dimension must be large enough to embed the dynamics of the underlying system.
Provided that this condition is satisfied, only the shape of the CMC function is influenced by the choice of the specific parameter value within a reasonable range, the conclusion about the direction of causal influence does not change.
The chosen embedding dimension determines the size of the embedding window, which introduces uncertainty in the temporal labeling of the reconstructed states.
Consequently, this causes uncertainty in the propagation delay of the causal effect, manifesting itself as a plateau in the CMC time-shift function.
This plateau complicates the precise determination of the effect propagation delay; hence, we recommend using the minimal embedding dimension that satisfies the necessary criteria.
Additionally, the embedding delay must be calibrated to the fastest frequency component in the signal, to mitigate the risk of aliasing.

A current weakness of CMC is that there is no statistical framework for significant link-detection backed by the theory of nonlinear dynamics.
This is a caveat compared to SGC and TES, where the statistical formulation and detection are already well-established \cite{bressler_2011}.
There were several previous attempts to create this statistical framework for topological causal discovery methods \cite{wang_improved_2019, krishna_inferring_2019, huang_systematic_2020, sarah_cobey_limits_2016, krakovska_testing_2016}, but these tools neglect significant information about the bundle structure of coupled dynamical systems \cite{stark_1999, benko_complete_2020}. Therefore, a proper statistical test is yet to be created.

Another interesting extension of CMC would be to detect time-varying connections by swapping simple coherence metrics for wavelet coherence. 
This would allow us to perform time-frequency analysis to investigate temporally varying frequency-specific causal relations between time series.

Finally, within the scope of this study, we introduced the CMC method to detect causality in the frequency domain using simulated datasets. Further research should include the application of the proposed method on diverse data, with the objective of elucidating novel insights into the underlying mechanisms of nonlinear dynamical systems.

\section{Conclusions}

    In this study, we proposed the Cross-Mapping Coherence (CMC) method for detecting frequency-dependent directional and circular causal interactions from time series data. We demonstrated the efficacy of CMC on various nonlinear model-systems, including logistic maps, Lorenz systems, Kuramoto oscillators, and the Wilson-Cowan model of V1-V4 cortical areas.

    The CMC method extends the Convergent Cross-Mapping method by changing performance evaluation from the coefficient of determination to coherence, and automates the detection of causal effect delay with peak prominence metrics.

    We demonstrated that CMC identifies complex causal interactions between nonlinear systems. CMC revealed uniform causality along each frequency band for coupled logistic maps, frequency-dependent interactions with higher power at lower frequencies for the Lorenz systems, and frequency-specific interactions in coupled Kuramoto oscillators. Moreover, it identified high frequency feedforward and low frequency feedback connections in the Wilson-Cowan model.

    We investigated how detection performance depends on sample size, coupling strength, and observation noise. In addition, we inspected the effects of the choice of embedding parameters and gave recommendations on how to choose them optimally.
    
    In summary, our findings reveal that the novel CMC method is efficient at revealing interactions between nonlinear systems in the frequency domain. Our results may have important implications for various fields, including neuroscience, Earth systems science, and engineering, where understanding complex nonlinear interactions is crucial.

\section{Methods}

    The simulations and analyses were carried out using the python numpy package\cite{harris2020array}, visualization were generated by matplotlib\cite{Hunter:2007}.

    \subsection{Data Simulations}

        \subsubsection{Logistic maps}
        Coupled logistic maps can be generated according to the equations:
        \begin{equation}
            x_i \left[t+1 \right] \;=\; \mathcal{F} \left( r_i \; \ x_i \left[t \right] \; \left(1 - \sum_jA_{ij} \; x_j \left[ t \right] \right) \right)
        \end{equation}
        where $r_i$ is the rate-parameter of the $i$th  logistic map, $A_{ij}$ is the coupling matrix with $A_{ii}=1$ and $\mathcal{F}$ is a mirroring boundary condition to constrain the data-point to the $[0,1]$ interval. 
        
        We simulated $n=10000$ datapoints with unidirectional, circular,common driver, and independent connection topologies. The simulation parameters are shown in Table\,\ref{tab:logmap_rates}.

        \begin{table}[htb!]
        	\centering
        	\begin{tabular}{c|c|c|c|c|c|c|c}
        		 scenario & $r_1$ & $r_2$ & $r_3$ & $A_{21}$ & $A_{12}$ & $A_{13}$ & $A_{23}$\\ 
        		\hline
        		$\rightarrow$ & $3.9902032398544094$ & $3.9900842430866197$ &  $0$ & $0.05$ & $0.0$ & $0.0$ & $0.0$ \\
        		$\leftrightarrow$ & $3.9903787118484475$ & $3.9900528401775186$ &  $0$ & $0.05$ & $0.05$ & $0.0$ & $0.0$ \\
        		$\Lsh \Rsh$ & $3.9903839016316964$ & $3.9904120926448896$ &  $3.9904110403001893$ & $0.0$ & $0.0$ & $0.05$ & $0.05$ \\
        		$\perp$ & $3.9903770705735107$ & $3.9907504255884914$ &  $0$ & $0.0$ & $0.0$ & $0.0$ & $0.0$ \\
        	\end{tabular}
        	\caption{The values of the rate  and coupling parameters of the coupled logistic maps.}
        	\label{tab:logmap_rates}
        \end{table}

        We tested the length dependence of CMC on unidirectionally coupled logistic map systems on $L=400$, $700$, $1000$, $2000$, $5000$ dataset lengths. The simulations were carried out with the rate parameters $r_1=3.99097965$ and $r_2=3.99024767$ and the coupling parameter was set to $A_{21} = 0.05$.

        The coupling strength sensitivity was checked on the same kind of coupled systems by varying the coupling strength parameter $A_{21}$ to take the values: $0$, $0.05$, $0.1$, $0.15$, and $0.2$ with rate parameters $r_1=r_2=3.99$ and data set length $L=2000$. 

        The robustness for observational noise was investigated in the signal-to-noise ratios of $1000$, $100$, $10$, $5$, and $2$ by adding Gaussian white noise to the signal with corresponding standard deviation. The rate parameters were $r_1=3.99097965$ and $r_2=3.99024767$, and the coupling parameter was set to $A_{21}=0.15$ with the data length as $L=2000$. 

        The effect of the embedding parameters on the inference results was analyzed by observing the shape of the CMC function depending on the embedding dimension in the range of [2, 10] with two-wise steps and with a data length of $L=20000$. The simulation parameters were the same as in the noise robustness analysis with the noise level set to zero.

        \subsubsection{Lorenz system}
        We simulated coupled Lorenz systems ($T=5000, \Delta t=10^{-3}$) according to the equations:
        
        \begin{equation}
        \begin{split}
            \dot{x}_1  &= \sigma_1 \left[(y_1 - x_1) + \kappa_1  (y_2 - x_1)\right] \\
            \dot{y}_1 &= x_1 (\rho_1 - z_1) - y_1 \\
            \dot{z}_1 &= x_1 y_1 - \beta_1 z_1\\
            \\
            \dot{x}_2 &= \sigma_2  \left[ (y_2 - x_2) + \kappa_2  (y_1 - x_2) \right] \\
            \dot{y}_2 &= x_2 (\rho_2 - z_2) - y_2\\
            \dot{z}_2 &= x_2  y_2 - \beta_2 z_2 \\
        \end{split}
        \end{equation}
        where parameters are signed as Greek lowercase letters, and the coupling is taking effect through the $\kappa_i$ terms. The parameter values are shown in Table\,\ref{tab:lorenz_params} and Table\,\ref{tab:lorenz_couple}. The integration was performed using the 4th order Runge-Kutta method with the scipy odeint function\cite{2020SciPy-NMeth}.

        \begin{table}[htb!]
            \centering
            \begin{tabular}{c|r}
                parameter & value \\
                \hline
        		$\sigma_{1}$ & $10.000$ \\
        		$\rho_{1}$ & $27.000$ \\
        		$\beta_{1}$ & $2.667$ \\
        		$\sigma_{2}$ & $10.209$ \\
        		$\rho_{2}$ & $25.900$ \\
        		$\beta_{2}$ & $2.652$ \\
        \end{tabular}
        \caption{fix parameter values for the Lorenz simulations.}
        \label{tab:lorenz_params}
        \end{table}

        \begin{table}[htb!]
            \centering
            \begin{tabular}{c|c|c|c}
                &  $\rightarrow$ &  $\leftrightarrow$ & $\perp$\\
                \hline
        		$\kappa_1$ & $0.0$ & $0.1$ & $0.0$\\
        		$\kappa_2$ & $0.1$ & $0.1$ & $0.0$\\
            \end{tabular}
            \caption{Coupling parameters for the Lorenz simulations}
            \label{tab:lorenz_couple}
        \end{table}
        
        \subsubsection{Kuramoto}
        We simulated a small network of Kuramoto oscillators to demonstrate directional couplings on different frequency bands with simulation step $\Delta t = 5 \times 10^{-3}$, $N=20000$:
        \begin{equation}
            \begin{split}
                    \dot{\theta}_i &= 2 \pi \nu_i + \frac{K_i}{m}\sum_{j=1}^m\sin(\theta_i - \theta_j) + \eta_i \\
                    x_i &= \sin\theta_i
            \end{split}
        \end{equation}
        where $m=3$, the basic frequencies are $\nu_i = [10.50422624, 59, 40]$ Hz, the coupling constants are $K_i=[0, 3, 4.3]$ , and $\eta_i \sim \mathcal{N}(0, 0.1)$ is a white noise term.
        The actual observed time series are the values of $x_i$. The numerical integration of the stochastic differential equation was carried out using the sdeint package\cite{aburn_critical_2017}.
        
        \subsubsection{Wilson-Cowan model}
        A nonlinear Wilson-Cowan-type neural mass model was employed to simulated the coupled cortical areas. The activity of a single area is described by four uniquely coupled stochastic differential equations, representing a pair of excitatory and inhibitory populations in two laminar compartments (supra- and infragranular) of the forms:

        \begin{equation}
            \tau_E\frac{dr_E}{dt} = -r_E + \Theta(I_E^{net} + I_E^{ext}) + \sqrt{\tau_E}\xi_E(t),
        \end{equation}
        
        \begin{equation}
            \tau_I\frac{dr_I}{dt} = -r_I + \Theta(I_I^{net} + I_I^{ext}) + \sqrt{\tau_I}\xi_I(t),
        \end{equation}
        where $r_{E,I}$ are dimensionless activities (i.e. mean firing rates) of an excitatory and an inhibitory population, respectively, $\tau_{E,I}$ are the corresponding time constants, $\xi_{E,I}$ are Gaussian noise terms with zero mean, and $\sigma_{E,I}$ strength and $\Theta(x) = x/(1-\exp^{-x})$ is the transduction function (used instead of a sigmoid function), while $I_{E,I}^{net}$ and $I_{E,I}^{ext}$ represent the input from other parts of the network and from exterior sources (e.g from the thalamus, which is not included as an individual node), respectively. See the supplementary materials of Mejias et al. (2016) \cite{mejias2016} for a detailed description of the original model, or Varga et al. (2021) \cite{varga_2021} for our specific implementation.

    \subsection{Data analysis steps}

        \subsubsection{Logistic maps}
        The logistic map data does not need any preprocessing; we embedded the time series and ran the CMC analysis in a $[-20, 20]$ time delay range with $E=2$, $\tau=1$ embedding parameters. We plotted the results on the shift-frequency axis (Fig.\,\ref{fig:logmap}). We used the same embedding parameter values for the data length, coupling sensitivity, and noise-robustness analyses.
        
        \subsubsection{Lorenz systems}
        For the Lorenz dataset, we subsampled the time series by a factor of $100$ and then applied the time delay embedding procedure on the $x_i$ variables from the subsystems ($E=7$, $tau=1$).
        We conducted the CMC analysis in the $[-2, 2]$ s range to discover time-delayed interactions (Fig.\,\ref{fig:lorenz}).

        \subsubsection{Kuramoto models}
        For the Kuramoto dataset, we chose the first 10000 points and ran the embedding procedure and subsequent CMC analysis ($E=5, \tau=1$) on the dataset with a $[-0.1, 0.1]$ s time-shift range (Fig.\,\ref{fig:kuramoto}).

        \subsubsection{Wilson-Cowan model}
        
        For the Wilson-Cowan dataset, we embedded the time series from V1 and V4 ($E=9, \tau=1$) and conducted the CMC analysis in a $[-0.2, 0.2]$ s time-shift range to detect time-delayed connections over the frequency bands.
        We carried out this analysis on $N=10$ instances and then averaged the results.
        We applied prominence calculations per frequency band on the causal side of the timeshift-frequency plot.
        Also, we averaged over the instances the per-frequency normalized results to transform out the effect of amplitude differences on different frequency bands and performed the prominence reduction technique.

\section{Acknowledgements}
    This research was founded by the Hungarian National Research, Development and Innovation Office K135837 (Z.S.) and the Hungarian Research Network, HUN-REN supported the project under the grant SA-114-2021 (Z.S.). 
\printbibliography
\end{document}